\documentclass{article}

\usepackage[final]{corl_2020} 

\usepackage{graphicx} 
\usepackage{subfig}
\usepackage{wrapfig}
\usepackage{bm}
\usepackage{amsmath}
\usepackage{amssymb}
\usepackage{mathrsfs}
\usepackage{booktabs}
\usepackage{multicol}
\usepackage{multirow}
\usepackage{longtable}
\usepackage{commath}
\usepackage{textcomp}
\usepackage{gensymb}
\usepackage{appendix}

\title{Deep Phase Correlation for End-to-End Heterogeneous Sensor Measurements Matching}

\author{
  Zexi Chen\;\;\;\;\;  Xuecheng Xu\;\;\;\;\;  Yue Wang$^{*}$\;\;\;\;\;  Rong Xiong\\
  Control Science and Egineering of Zhejiang University\\
  \texttt{\{chenzexi,xuechengxu,ywang24,rxiong\}@zju.edu.cn} \\
  $^{*}$ is the corresponding author
}

\begin{document}
\maketitle

\vspace{-0.8cm}
\begin{abstract}
    The crucial step for localization is to match the current observation to the map. When the two sensor modalities are significantly different, matching becomes challenging. In this paper, we present an end-to-end deep phase correlation network (DPCN) to match heterogeneous sensor measurements. In DPCN, the primary component is a differentiable correlation-based estimator that back-propagates the pose error to learnable feature extractors, which addresses the problem that there are no direct common features for supervision. Also, it eliminates the exhaustive evaluation in some previous methods, improving efficiency. With the interpretable modeling, the network is light-weighted and promising for better generalization. We evaluate the system on both the simulation data and Aero-Ground Dataset which consists of heterogeneous sensor images and aerial images acquired by satellites or aerial robots. The results show that our method is able to match the heterogeneous sensor measurements, outperforming the comparative traditional phase correlation and other learning-based methods. {\small \href{https://github.com/jessychen1016/DPCN}{\texttt{Code is available here}}.}

\end{abstract}



\section{Introduction}

\begin{figure}[ht]
    \centering
        \includegraphics[width=\linewidth]{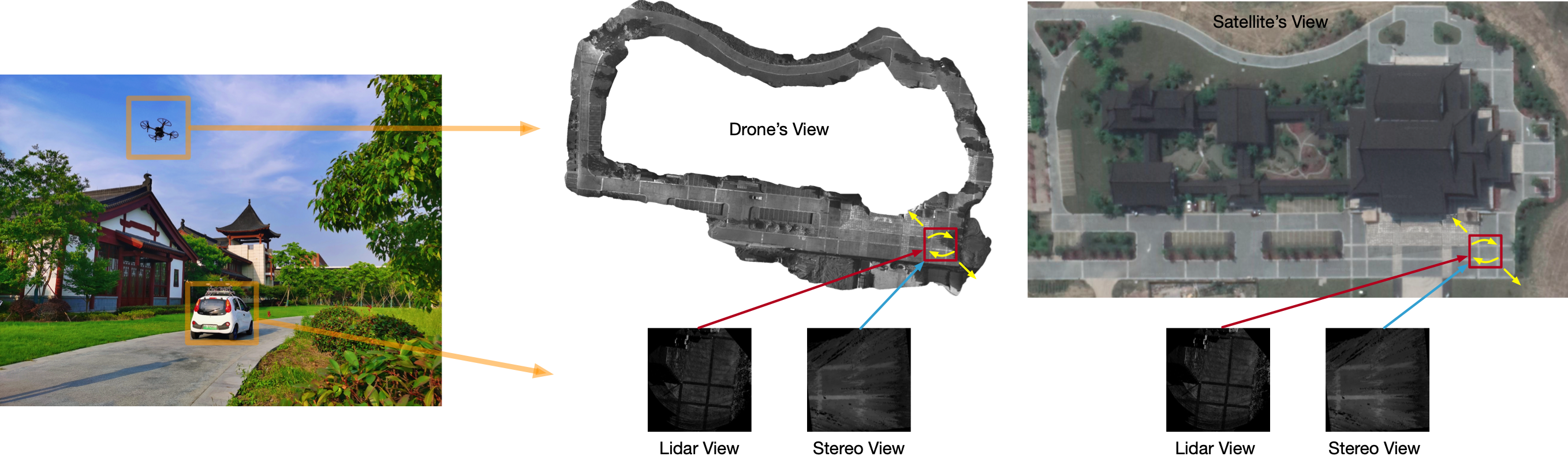}
    \caption{A typical scenario for localization based on matching measurements from heterogeneous sensors. \textbf{Left}: Aerial-Ground cooperation; \textbf{middle}: matching observations from LiDAR or stereo cameras of the ground robot to drone's global map; \textbf{right}: matching observations from LiDAR or stereo cameras of the ground robot to satellite's global map.}
    \label{fig:instant of heterogeneous}
    \vspace{-0cm}
\end{figure}

    Localization is one of the most fundamental problem for mobile robots. With a decade of research, localization given the measurement and the map built by the same sensor is relatively mature. But for matching measurements from heterogeneous sensor modalities remains an open problem. This problem is practical considering the effort to build a map. We would like the map to be sharable by multiple robot users, even equipped with heterogeneous sensors. Note the great progress in visual inertial navigation and mapping \cite{Mourikis2007}, the solution space to localization can be reduced to 3 or 4, namely the translation and the heading angle. By exploiting birds-eye view measurements, the problem is re-formulated as an image matching problem with warping space built on $\mathbb{SIM}(2)$. In this paper, we focus on this scenarios with birds-eye view images acquired by ground vehicles, satellites and UAV with vision and LiDAR, as shown in Figure \ref{fig:instant of heterogeneous}.

    Previous researches on homogeneous image matching can be divided into two categories: those rely on point features correspondences to localize in specific setups \cite{Noda2011, Lin2015, Barnes2017, Cheng2019}, those apply dense correlation methods to find the best pose candidate in solution space \cite{SrinivasaReddy1996, Kaslin2016, Talmi2017, Kim2018, Barnes2019}. However, all these approaches do not perform well when heterogeneous measurements are given. As for heterogeneous image matching, \cite{Kummerle2010, Ruchti2015} utilize hand-craft features to localize LiDAR against satellite maps. These frameworks rely heavily on the coverage and optimality of the hand-crafted features over the variations in real environment. Learning-based methods proved to have certain generalization ability are intuitively appropriate for heterogeneous image matching. \cite{Lu2019, Tang2020} learn the embeddings for heterogeneous observations and exhaustively search for the optimal pose in the discrete solution space, which is more interpretable than regressing the pose by an end-to-end network as in \cite{Kendall2015, Kendall2017}. However, the former suffers from low efficiency and limited pose range due to the exhaustive evaluation on the large pose space, constraining the application with known scale.

    By revealing the problems encountered by prior works, one can think of an ideal matcher that can obtain the solution without exhaustive evaluation and also have good interpretability and generalization. In our work, we set to propose such a learnable matcher, of which the essence is a differentiable phase correlation. Phase correlation \cite{SrinivasaReddy1996} is a similarity-based matcher that performs well for inputs with the same modality but only tolerate small high frequency noise. We modify the phase correlation into a differentiable manner and embed it into our end-to-end framework. This architecture allows our system to find optimal feature extractors with respect to the resultant pose of image matching. Specifically, we adopt the conventional phase correlation pipeline proposed by \cite{SrinivasaReddy1996} and explicitly endow the Discrete Fourier Transform (DFT) layer, log-polar transformation layer (LPT), and differentiable correlation layer (DC) with differentiability and thus make it trainable for our end-to-end matching network as shown in Figure \ref{fig:demo net}. Our experiments show the robustness and efficiency of the propose method on matching heterogeneous sensor measurements.
\vspace{-0.2cm}
\section{Deep Phase Correlation Network}
\label{sec:LogPolarTemplateMatchingNetworkwithDomainTransferring}

    With known gravity direction, the relative pose between the two observed birds-eye view images $ (O_{1},O_{2}) $ can be simplified as similarity transform $S$:
\begin{equation}
    S = \begin{pmatrix}
    s\boldsymbol{R}_{\theta} & \boldsymbol{t} \\
    0 & 1
    \end{pmatrix} \in \mathbb{SIM}(2)
    \label{eq: pose}
\end{equation}
    where $s \in \mathbb{R}^+$ is the scale, $\boldsymbol{R}_{\theta} \in \mathbb{SO}(2)$ is the rotation matrix generated by the heading angle $\theta$, and $\boldsymbol{t} \in \mathbb{R}^2$ is the translation vector. In general, the two images are disturbed by illumination, shadow and occlusion, or even acquired by heterogeneous sensors. For example, $O_{p1}$ is acquired by a birds-eye camera of a UAV in the morning, while $O_{p2}$ is a local elevation map constructed by the ground robot with a LiDAR. To address this issue, a general process is to extract features from the two images, and estimate the relative pose using the features instead of the original sensor measurements. By applying the template matching to the features, we derive the optimal feature extractors $f_{1}$ and $f_{2}$ for the two images respectively by solving
    \begin{equation}
    \min_{f_1,f_2} \| S^*-\arg\max_{S} H( f_{1}(O_{1}),T\{ f_{2}(O_{2}), S \}) \|  
    \label{eq: undifferentiable pipeline}
    \end{equation}
    where $S^*$ is the ground truth, $H$ is a scoring function to measure the similarity, and $T\{\cdot,S\}$ is to transform $\cdot$ by relative pose $S$. Inner product is often regarded as the scorer function. Then we have
    \begin{equation}
    \min_{f_1,f_2} \| S^*-\arg\max_{S} f_{1}(O_{1}) \cdot T\{ f_{2}(O_{2}), S \} \|
    \label{eq: corr}
    \end{equation}
    There are two problems. First, an exhaustive evaluation is required for all elements in $S$, which is extremely time consuming considering the $4$ dimensional space $\mathbb{SIM}(2)$. Second, $\arg\max$ is not differentiable, thus (\ref{eq: corr}) is hard to be optimized. If $f_1$ and $f_2$ are hand-crafted processes, the optimization is even harder. In this paper, we set to differentiate (\ref{eq: corr}) and eliminate the exhaustive evaluation to find efficient data-driven feature extractors $f_1$ and $f_2$.

    \subsection{Decoupled correlation based pose estimator}
    \label{subsec: DFT&LPT}

    \noindent{\textbf{Cross-correlation:}} We begin with the known scale and rotation $\{\theta^*,s^*\}$, reducing the unknown parameters to $2$. Denoting $F_1 = f_1(O_1)$ and $F_2 = T\{f_2(O_2), \{\theta^*,s^*\} \}$, we have
    \begin{equation}
    \min_{f_1,f_2} \| \boldsymbol{t}^*-\arg\max_{\boldsymbol{t}} \sum_{x} F_{1}(x) F_{2}(x+\boldsymbol{t}) \|
    \label{eq: corrt}
    \end{equation}
    where $x$ is a position in the feature. Note the term in the $\arg\max$ is the cross-correlation function parameterized by $\boldsymbol{t}$ between $F_1$ and $F_2$, which can be evaluated very efficiently using convolution.

    \noindent{\textbf{Translation invariance:}} In general, if we achieve scale and rotation by assuming known translation, then the problem is chicken-and-egg. We introduce a representation which is invariant to translation, but variant to scale and rotation. Thus we can ignore the translation when solving for scale and rotation. Specifically, we refer to the magnitude of frequency spectrum $\Omega$ of $O_1$ and $O_2$. According to the property of Fourier transform, we have
    \begin{equation}
    |\mathfrak{F}\{T\{f_2(O_2),  \{\theta,s\} \}\}| = |\mathfrak{F}\{T\{f_2(O_2), S \}\}|
    \label{eq: fft}
    \end{equation}
    where $\mathfrak{F}$ is the fourier transform. It means that only rotation and scale have effects on the magnitude of frequency spectrum.
    \begin{equation}
    \min_{f_1,f_2} \| \{\theta^*,s^*\}-\arg\max_{\theta,s} \Omega_1 \cdot \Omega_2(\{\theta,s\}) \} \|
    \label{eq: onlyrs}
    \end{equation}

    \noindent{\textbf{Cartesian to log-polar:}} We decouple rotation and scale from relative pose using the frequency spectrum. However, exhaustive evaluation is still required for rotation and scale. By looking into the frequency domain, we have
    \begin{equation}
    \Omega_1(\psi) = \Omega_2(s^*\boldsymbol{R}_{\theta^*}\psi)
    \label{eq: onlyrcart}
    \end{equation}
    where $\psi$ is a position in the frequency spectrum. By representing it in the polar coordinates, we have
    \begin{equation}
    \Omega_1(\alpha,\beta) = \Omega_2(s^*\alpha,\beta+\theta^*)
    \label{eq: onlyrpolar}
    \end{equation}
    where $\alpha = |\psi|$ and $\beta = \angle \psi$. To deal with the scale, we further apply the logarithm to the position
    \begin{equation}
    \Omega_1(\log\alpha,\beta) = \Omega_2(\log\alpha+\log s^*,\beta+\theta^*)
    \label{eq: onlyrlogpolar}
    \end{equation}
    Now we finally arrive at the correlation form with respect to the scale and rotation, which eliminates all exhaustive evaluation in the whole pose estimator.
    \begin{equation}
    \min_{f_1,f_2} \| \{\log s^*,\theta^*\}-\arg\max_{\{\log s,\theta\}} \sum_{\{\log\alpha,\beta\}} \Omega_1(\log\alpha,\beta) \Omega_2(\log\alpha+\log s,\beta+\theta) \|
    \label{eq: corrrs}
    \end{equation}
    \vspace{-0.4cm}

    When there is no feature extractor, or equivalently, the original sensor measurements are fed, the process is called phase correlation, estimating the relative pose in $\mathbb{SIM}(2)$ very efficiently. However, when there is variation between the pair of inputs, the feature extractor is indispensable.

    \subsection{End-to-end learnable feature extractor}
    \label{subsec: End-to-end learnable feature extractor}

\begin{figure}[t]

    \centering
    \includegraphics[width=\linewidth]{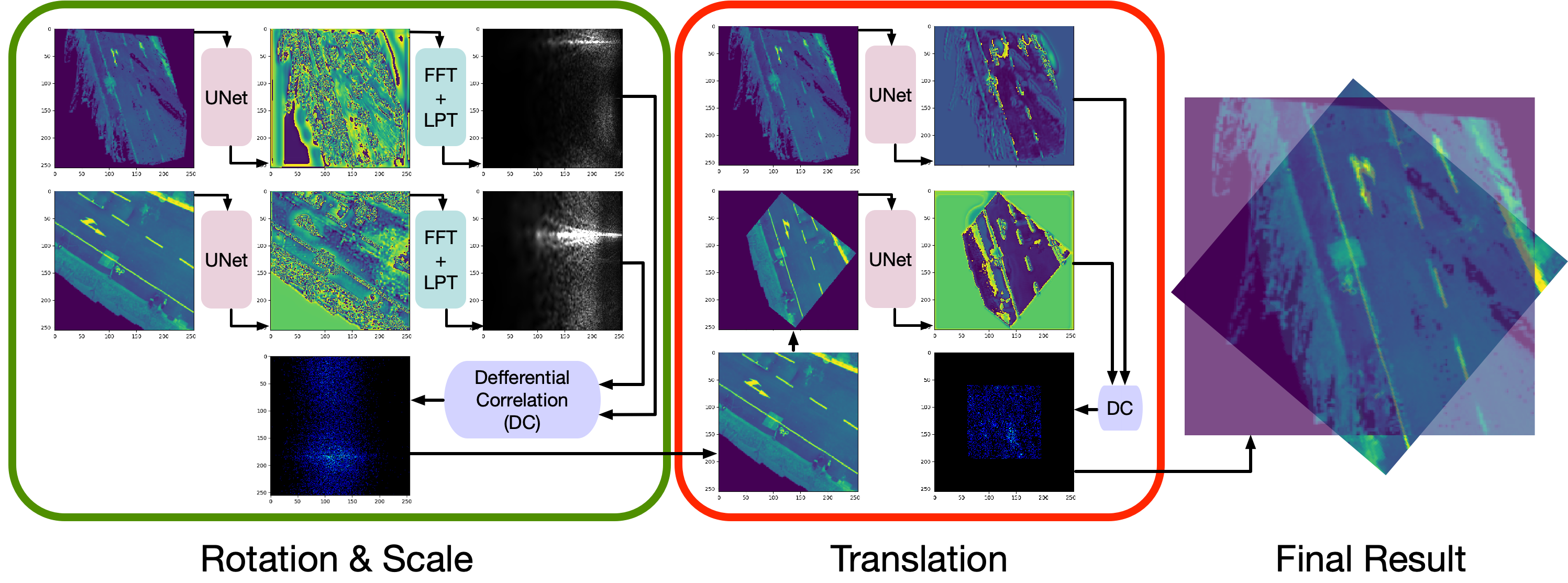}
    \caption{The architecture of proposed Deep Phase Correlation Network (DPCN), which consists of 4 U-Net based feature extractors and differentiable phase correlation for decoupled scale, rotation and translation estimation.}
    \label{fig:demo net}
    \vspace{-0.1cm}
\end{figure}

    \noindent{\textbf{Expectation as differentiable $\arg\max$:}} To find the feature extractor which is optimal with respect to the training data, we have to approximate the $\arg\max$ in (\ref{eq: corrt}) and (\ref{eq: corrrs}) with differentiable function. We connect the cross-correlation to a softmax function, which maps the input to a discrete probability density function $p(\cdot)$. Set translation part as example, the cross-correlation in (\ref{eq: corrt}) is
    \begin{equation}\label{approxexp}
      p(\boldsymbol{t})=\frac{\exp(-\sum_{x} F_{1}(x) F_{2}(x+\boldsymbol{t}))}{\sum_{\boldsymbol{t}}\exp(-\sum_{x} F_{1}(x) F_{2}(x+\boldsymbol{t}))}
    \end{equation}
    By keeping the features positive, we do not need to care about the negative correlation. Still refering to the translation part, given the probability function, we derive the expectation as the optimal translation estimator, and substitute it into (\ref{eq: corrt}), yielding the differentiable form
    \begin{equation}\label{exp}
    \min_{f_1,f_2} \| \boldsymbol{t}^*- \sum_{\boldsymbol{t}} \boldsymbol{t} p(\boldsymbol{t}) \|
    \end{equation}
    For rotation and scale (\ref{eq: corrrs}), the same expectation estimator can be applied to approximate the non-differentiable $\arg\max$ based estimator. When learning, we assign a temperature coefficient to the softmax to tune the range of feature input, which accelerates the convergence, but does not make difference in theoretic derivation. The whole pose estimator can be regarded as a differentiable phase correlation (DPC) with back-propagated gradients to enforce the learning of feature extractor.

    \noindent{\textbf{Deep feature extractor:}} The conventional phase correlation \cite{SrinivasaReddy1996} utilizes high pass filters to suppress high frequency random noise of two inputs, which can be seen as a feature extractior. For more distinct variation between the pair of inputs, one high pass filter is far from sufficient. Considering that there is no common feature to directly supervise the feature extractor, we utilize the end-to-end learning to address the problem.

    We adopt U-Net for feature extraction, aiming at learning the common features of the two images implicitly. We construct 4 separate U-Nets with the input and output size of $256\times256$ respectively for the template image and the source image in the rotation phase and the translation phase, shown in Figure \ref{fig:demo net}. Each U-Net is constructed with 4 down-sampling encoder layers and 4 up-sampling decoder layers to extract features. As the training progresses, the parameters of four U-Nets are tuned. Note that this network is light-weighted so that it could be efficient enough for real-time execution. Combining the feature extractor and the DPC, we name the whole network as DPCN.

\subsection{Data preparation for learning}
\label{subsec:PhaseCorr}


    One question is that why we supervise DPCN on the estimated pose, but not the correlation matrix with a one-peak matrix centering at the correct position, thus the expectation based estimator can be eliminated. We argue that enforcing the correlation matrix to be one-peak is over-supervision. In theory of phase correlation, the position corresponding to the maximum correlation, does not necessarily mean zero correlation for the others, which can be explained by the resonance in physics. However, when the correlation map is passed through softmax estimator, the temperature coefficient suppresses the non-maximal part in the correlation matrix to be close to 0, resulting in a normalized probability density function. In addition to the pose supervision as shown in (\ref{exp}), we also supervise the probability density function of translation, and scale/rotation, to be one-peak. Still refer to the translation as example, we do it by applying the KL-divergence:
    \begin{equation}\label{kld}
    \min_{f_1,f_2} \delta_{\boldsymbol{t}^*}(t) p(\boldsymbol{t})
    \end{equation}
    where $\delta_{\boldsymbol{t}^*}(t)$ is a normalized one-peak function centering at $\boldsymbol{t}^*$. In practice, we slightly expand the one-peak function with Gaussian smooth. Theoretically, the resultant distribution for some cases can be multi-modal, e.g. the repetitive local environment. However, given massive training data with similar input and different output, the optimal solution to (\ref{kld}) is a multi-modal probability function, i.e. multi-modal $ p(\boldsymbol{t})$. The two loss (\ref{kld}) and (\ref{exp}) are mixed by weight. For translation part, multi-modal result occurs more often, so we increase the loss of (\ref{kld}) in the total loss for translation phase. 

\section{Experimental Results}
\label{sec:result}
    In this section, we explicitly evaluate the performance of our approach. By utilizing phase correlation with DFT and log-polar transform over the learned representation of two images, we are able to fully estimate the rotation and scale transformation of the two heterogeneous images, and eventually able to estimate the 4-DoF relative pose $S$.

\begin{figure}[t]
    \centering
    \includegraphics[width=\linewidth]{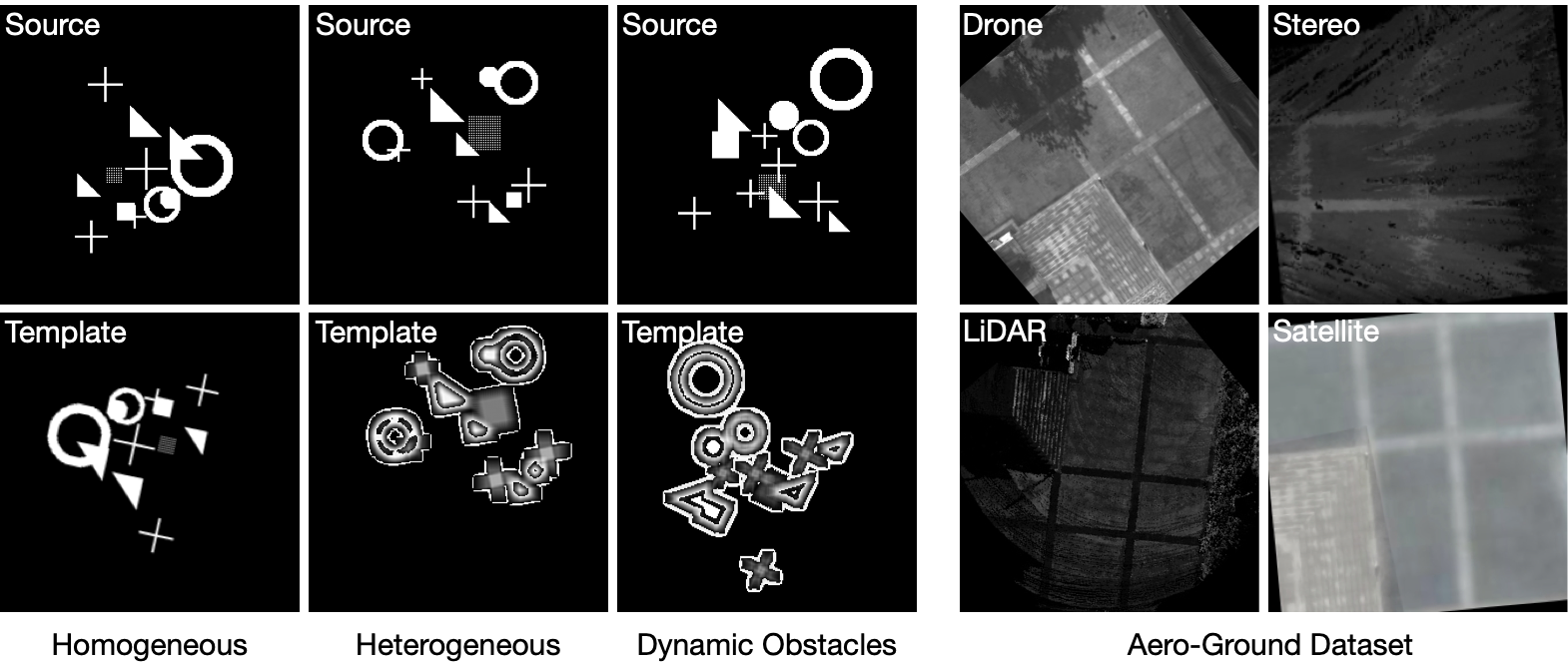}
    \caption{\textbf{Simulation dataset(left)} containing ``Homogeneous", ``Heterogeneous" and ``Dynamic Obstacles" sets. \textbf{Aero-Ground Dataset(right)} containing ``drone's view", ``LiDAR intensity", ``stereo" and ``satellite"}
    \label{fig:dataset}
    \vspace{-0.4cm}
\end{figure}

\begin{figure}[t]
\includegraphics[width=\linewidth]{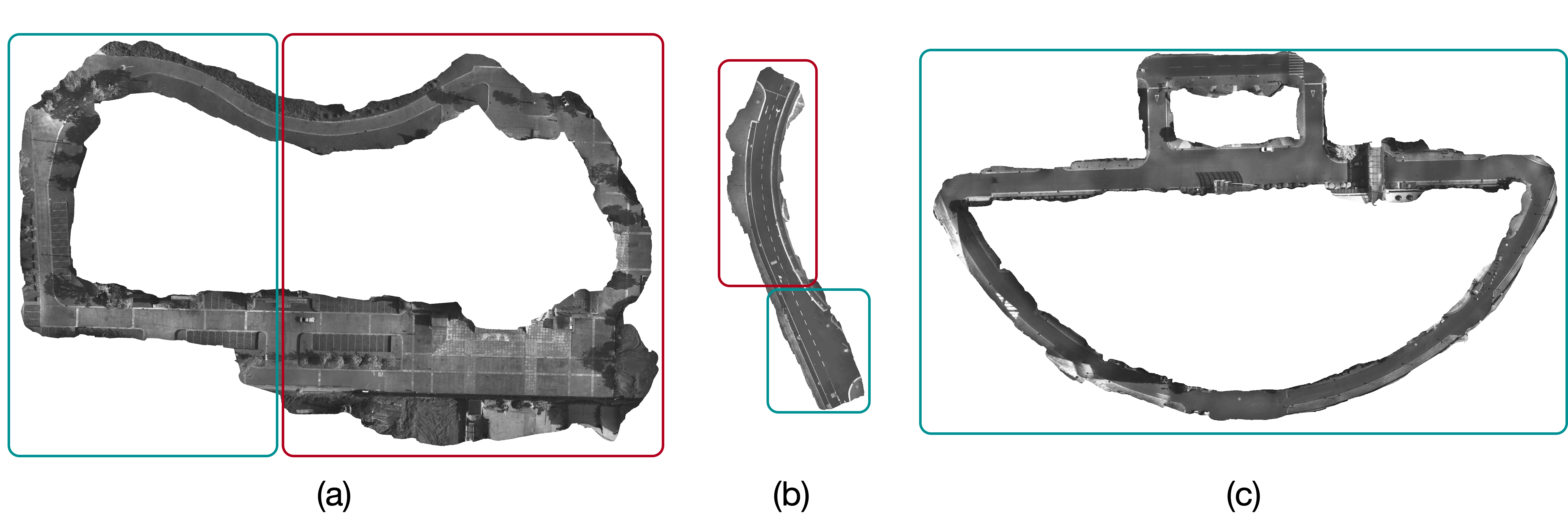}
\caption{Demonstration of birds-eye views of the different scenes in Aero-Ground dataset. The experiments are carried out on location (a) and (b) separately in which the model is trained on images pairs generated inside \textbf{red} areas and validated on images pairs generated inside \textbf{blue} area. The generalization experiment is carried out with estimating poses of images inside location (c) with models trained on (a) and (b).}
\label{fig:AG Dataset}
\vspace{-0.4cm}
\end{figure}

   \textbf{\textsc{Dataset} \& Metrics: }Our approach is evaluated both on randomly generated simulation dataset and on real-world dataset "Aero-Ground Dataset"\cite{AeroGroundDataset} which contains several different image pairs shown as follows:
\begin{itemize}
    \vspace{-8pt}
    \item \textbf{l2d:}``LiDAR Local Map" to ``Drone's Birds-eye Camera View";
    \vspace{-3pt}
    \item \textbf{l2sat:}``LiDAR Local Map" to ``Satellite Map";
    \vspace{-3pt}
    \item \textbf{s2d:}``Stereo Local Map" to ``Drone's Birds-eye Camera View";
    \vspace{-3pt}
    \item \textbf{s2sat:}``Stereo Local Map" to ``Satellite Map".
    \vspace{-8pt}

\end{itemize}
    On the simulation dataset, we evaluate our work on homogeneous, heterogeneous, and heterogeneous with dynamic obstacles images pairs whereas on the Aero-Ground Dataset, we evaluate our work on the application of cooperative SLAM system between ground mobile robots, the MAV and the Satellite. The demonstration of both datasets is shown in Figure \ref{fig:dataset} and \ref{fig:AG Dataset}. Finally, we applied our method to Monte Carlo Localization to prove the real-time capability and robustness of our method. In all datasets, we constrained translations of both $x$ and $y$, rotation changes and scale changes in the range of $[-50, 50]$ pixels, $[0, \pi)$ and $[0.8, 1.2]$ respectively with images shapes of $256 \times 256$.

    For evaluating the accuracy and  error rate of estimation, we consider ``Accuracy in Units" and Mean Square Error(MSE) of the estimated result and the ground truth as the mean indicator:

\vspace{-0.5cm}
\begin{equation}
     Acc_{v_t} = \frac{1}{N} \cdot \sum\nolimits_{i=1}^N\textbf{Count}\{\abs{S_i-G_i} \leqslant t\} \times 100\% 
\label{eq: Acc}
\end{equation}

\vspace{-0.4cm}
\begin{equation}
     E = \frac{1}{N} \cdot \sum\nolimits_{i=1}^N(S_i-G_i)^2
\label{eq: Err}
\end{equation}
    where $v$ being the output type(x, y, rotation, and scale), $t$ being the threshold of accuracy (pixel for translation, degree for rotation and multiplier for scale), $N$ being the amount of image pairs, $S_i$ being the estimated result of the $i$th image pair and $G_i$ being the corresponding ground truth. ``Accuracy in Units" is calculated as the percentage of estimation with an error lower than the threshold.

    \textbf{\textsc{Comparative Methods: }}Benchmarks in the experiments include conventional \textsc{Phase Correlation}\cite{SrinivasaReddy1996}, deep learning based \textsc{QATM}\cite{Cheng2019}, \textsc{DAM}\cite{Park2020}, \textsc{Relative Pose Regression}(RPR), and \textsc{Dense Search}(DS). \textbf{\textsc{Phase Correlation}} is the baseline for registering two homogeneous images and the pipeline of which is also partly adopted in our approach. We select it as a benchmark for evaluating the performance of our approach in matching homogeneous images. \textbf{\textsc{QATM}} is a representative work in image matching applying deep learning to learn features for matching and NMS for matching selection. It could handles translation displacement with high accuracy and therefore we select it as the benchmark for evaluating translation estimations in heterogeneous images. Unfortunately, the author of QATM only provided a pretrained model without a training script so that we could only evaluate its performance with the provided model. \textbf{\textsc{DAM}} trains and learns affine transformations including translation, rotation and scale changes, and therefore we trained DAM on the same dataset and compare our method with it on the four aspects.  \textbf{\textsc{Relative Pose Regression}(RPR)} is adopted by multiple methods\cite{Kendall2015,Kendall2017} in pose regression by outputting desired classes of estimation through several fully-connected layers without an analytical solutions. Finally, \textbf{\textsc{Dense Search}(DS)} is the methodology adopted by \cite{Barnes2019} whcih violently rotates images to some discrete angles to estimate the relative shifts. We trained DS in a smaller rotation range [0,15] due to its discrete property and time consumption. It is selected as a benchmark to demonstrate the advantage of estimation continuity and swiftness of our approach.

    \subsection{Simulation Dataset}
        \label{subsec:SimulationDataset}
    In order to verify the feasibility of our approach under several different situations and the generalization capability of the fully differentiable methodology, we conduct various of experiments on the simulation dataset. The experiments on a pair of homogeneous images is conducted to verify the equivalence to the conventional phase correlation when dealing with images of same styles. Moreover, with experiments on a pair of heterogeneous images conducted, we show the unique capability of our approach with estimating the 4-DoF pose $S$ of two images with drastic style changes. Finally, we introduce dynamic obstacles to the heterogeneous image pairs to demonstrate the robustness of our approach for being insensitive to dynamic obstacles.

\begin{table}[ht]
\vspace{-0.4cm}
\captionsetup{justification=raggedright, singlelinecheck=true}
\caption{Results of simulation dataset. We choose the threshold error of $5 \:pixels$ for translation, $1\degree$ for rotation and $0.2\times$ for scale. More thresholds are elaborated in Appendix \ref{appendix: Elaboration on Translation Estimation}}
\label{results: simulation}
\centering
\resizebox{\textwidth}{!}{
\begin{tabular}{@{}cllllllllll@{}}
\toprule[1pt]
\multicolumn{1}{l}{\textbf{Benchmarks}} &
  Exp. &
  $E_x$ &
  $Acc_{x_5}\%$ &
  $E_y$ &
  $Acc_{y_5}\%$ &
  $E_r$ &
  $Acc_{r_1}\%$ &
  $E_s$ &
  $Acc_{s_{0.2}}\%$ &
  Runtime(ms) \\ \midrule
\begin{tabular}[c]{@{}c@{}}\textsc{\underline{Image}}\\ \textsc{\underline{Registration}}\end{tabular}
                     & 1 & \textcolor{blue}{0.6635} & \textcolor{blue}{99.2} & \textcolor{blue}{0.9231} & \textcolor{blue}{99.5} & \textcolor{blue}{0.0663} & \textcolor{blue}{99.7} & \textcolor{blue}{0.0710} & \textcolor{blue}{98.9} & 141.4 \\
                     & 2 & 1774.1592 & 52.3 & 3233.8133 & 33.2 & 145.8561 & 72.3 & 0.1992 & \textcolor{blue}{97.6} & 138.8 \\
                     & 3 & 2319.5537 & 49.2 & 2945.3017 & 42.6 & 121.5026 & 67.9 & \textcolor{blue}{0.1218} & \textcolor{blue}{96.7} & 137.0 \\
\underline{\textsc{QATM}}        & 1 & 15.4820 & 95.4 & 8.9192 & 96.3 & $\setminus$ & $\setminus$ & $\setminus$ & $\setminus$ & 108.3 \\
                     & 2 & 2999.3710 & 31.4 & 4286.4810 & 26.4 & $\setminus$ & $\setminus$ & $\setminus$ & $\setminus$ & 108.9 \\
                     & 3 & 3651.4691 & 25.6 & 4901.7201 & 21.5 & $\setminus$ & $\setminus$ & $\setminus$ & $\setminus$ & 109.1 \\
\underline{\textsc{DAM }}        & 1 & 53.7597 & 90.6 & 28.6825 & 95.9 & 19.2243 & 81.7 & 0.1452 & 90.5 & 111.7 \\
                     & 2 & \textcolor{blue}{2.5816} & \textcolor{blue}{98.4} & \textcolor{blue}{4.6234} & \textcolor{blue}{97.9} & \textcolor{blue}{28.9341} & \textcolor{blue}{80.8} & 0.1432 & 90.9 & 114.2 \\
                     & 3 & 46.1165 & 71.3 & 89.6835 & 68.2 & \textcolor{blue}{36.5608} & \textcolor{blue}{77.8} & 0.3625 & 87.3 & 110.4 \\
\underline{\textsc{RPR}}         & 1 & 8.6754 & 96.9 & 2.2201 & 97.3 & 16.2723 & 90.2 & 0.0805 & 95.2 & \textcolor{red}{64.1} \\
                     & 2 & 22.3634 & 39.9 & 32.8334 & 41.2 & 97.8517 & 78.3 & \textcolor{blue}{0.1367} & 96.7 & \textcolor{red}{64.7} \\
                     & 3 & 14.6842 & 51.1 & 11.1322 & 56.8 & 101.3329 & 76.8 & 0.1846 & 96.1 & \textcolor{red}{63.5} \\
\underline{\textsc{DS}}          & 1 & 7.4092 & 85.1 & 11.1108 & 79.3 & 26.2656 & 33.4 & $\setminus$ & $\setminus$ & 304.5 \\
                     & 2 & 5.4970 & 86.6 & 7.3333 & 86.5 & 31.3891 & 23.9 & $\setminus$ & $\setminus$ & 301.3 \\
                     & 3 & \textcolor{blue}{6.2850} & \textcolor{blue}{83.9} & \textcolor{blue}{7.8176} & \textcolor{blue}{79.6} & 25.6643 & 27.1 & $\setminus$ & $\setminus$ & 301.4 \\
\textbf{\textsc{\underline{DPCN}}} & 1 & \textcolor{red}{0.1031} & \textcolor{red}{100} & \textcolor{red}{0.2162} & \textcolor{red}{100} & \textcolor{red}{0.0528} & \textcolor{red}{100} & \textcolor{red}{0.0522} & \textcolor{red}{100} & \textcolor{blue}{71.9}\\
\textbf{\underline{(Ours)}}         & 2 & \textcolor{red}{0.0073} & \textcolor{red}{100} & \textcolor{red}{0.0172} & \textcolor{red}{100} & \textcolor{red}{0.0397} & \textcolor{red}{100} & \textcolor{red}{0.0642} & \textcolor{red}{100} & \textcolor{blue}{72.1}\\
                     & 3 & \textcolor{red}{0.0761} & \textcolor{red}{100} & \textcolor{red}{0.4671} & \textcolor{red}{100} & \textcolor{red}{0.0913} & \textcolor{red}{100} & \textcolor{red}{0.0013} & \textcolor{red}{100} & \textcolor{blue}{72.1}\\
\bottomrule[1pt]
\end{tabular}
}
Note:  $Exp.1$, $Exp.2$ and $Exp.3$ is conducted on ``Homogeneous", ``Heterogeneous", and ``Dynamic Obstacles" sets respectively. \textcolor{red}{Red} is the best performance and \textcolor{blue}{blue} is the secondary.
\vspace{-0.2cm}
\end{table}

    Experimental results on the simulation dataset are shown in Table \ref{results: simulation}. The results show that for the homogeneous pairs of images, our method maintains an equivalent performance with the conventional phase correlation pipeline in accuracy with faster speed and outperformed the rest of the baselines. In the heterogeneous dataset and dynamic obstacle dataset, our approach outperformed any existing baselines in accuracy and is only a bit slower than the Relative Pose Regression.

    The key to our approach is applying a fully optimal differentiable DFT with log-polar transformation to back-propagate the supervised error to train a multi-channel feature-based representation which is optimized for phase correlation to estimate the relative pose. It outperformed the conventional phase correlation method applying high pass filter with the learned representation when dealing with images that is not simply a cutout to one another. The learned representation, the result of the DFT, log-polar transformation, and the correlation map shown in Figure 6 indicate that the network is interpretable: by supervising end-to-end, the network is finally able to predict the transformation of two heterogeneous images with their U-Net outputs of convergent features. Note that the simulation dataset is randomly generated to reduce overfitting.

\begin{figure}[ht]
\centering
\includegraphics[width=\linewidth]{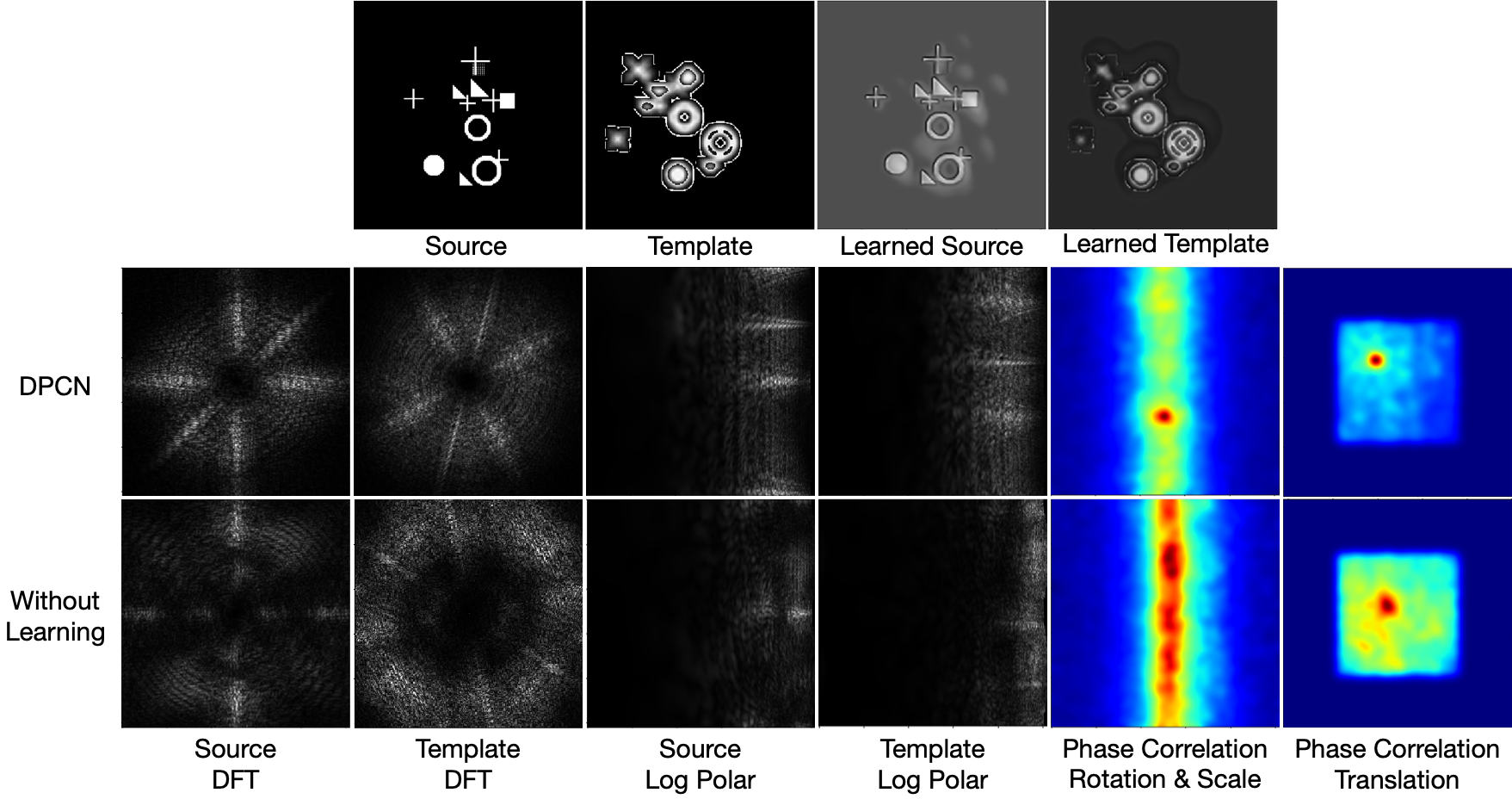}
\vspace{-0.4cm}
\caption{Workflow visualization showing that with codecs applied and trained, the network is able to reduce the impact of noise and style changes while conventional phase correlation without feature learning fails to recognize the relations of two images. }
\label{fig:simulation workflow}
\vspace{-0cm}
\end{figure}

    Table \ref{results: generalizationSimulation} and elaborations in Appendix \ref{appendix: Elaboration on Translation Estimation} verify the generalizing ability of our approach For generalization experiments, the model is trained on the ``Heterogeneous" set and is evaluated on two other sets in simulation datasets. The results show that with models not specifically trained, it can still maintain a high rate of accuracy in all 4-DoF.


\begin{table}[h]
\vspace{-0.4cm}
\renewcommand\arraystretch{1.4}
\captionsetup{justification=raggedright}
\caption{Results of generalization experiments with simulation dataset.}
\begin{small}
\resizebox{\textwidth}{!}{
\begin{tabular}{c ccccccccc ccccccccc }
\toprule[1pt]
Exp. & $E_x$ & $Acc_{x_5}(\%)$ & $E_y$ & $Acc_{y_5}(\%)$ & $E_{rot}$ & $Acc_{rot_{1}}(\%)$ & $E_{scale}$ & $Acc_{scale_{0.2}}(\%)$ \\ \midrule
Homogeneous       &  0.4121  &  100   &  0.705   &  100   &  0.0432  &  100   &  0.015  & 100\\
Dynamic Obstacle  &  0.0276  & 100   &  0.105   &   100  &  0.039  &  100   &  0.003   &  100\\
\bottomrule[1pt]
\end{tabular}
}
\end{small}
\label{results: generalizationSimulation}
\vspace{-0.3cm}
\end{table}

\subsection{Aero-Ground Dataset}
    \label{subsec:Aero-GroundDataset}

    To evaluate the applicability in the real world of our approach, we conduct experiments on Aero-Ground Dataset to match images across sensors. Few baseline support 4-DoF(or more) pose estimations across sensors and therefore we relax the condition by providing rotation and scale ground truth as initials to several benchmark for comparison, including the DAM with scaling initialized and QATM with both rotation and scale initialized. We train and validate our models in two different scenes (Figure \ref{fig:AG Dataset}(a) and (b)) separately and generalize them in the third scene(Figure \ref{fig:AG Dataset}(c)).

\begin{table}[ht]
\vspace{-0.4cm}
\caption{Results of the scene (a) and (b). More thresholds are elaborated in Appendix \ref{appendix: Elaboration on Translation Estimation}}
\label{results: Aero-Ground Dataset}
\resizebox{\textwidth}{!}{
\begin{tabular}{@{}lcllllllllll@{}}
\toprule[1pt]
Scene & \multicolumn{1}{l}{\textbf{Benchmarks}} & Exp. & $E_x$ & $Acc_{x_{10}}\%)$ & $E_{y}$ & $Acc_{y_{10}}(\%)$ & $E_{rot}$ & $Acc_{rot_{1}}(\%)$ & $E_{scale}$ & $Acc_{scale_{0.2}}(\%)$ & Runtime(ms) \\ \midrule
\multirow{13}{*}{(a)} & \underline{QATM}    & l2sat & 6533.8164 & 21.5 & 4082.8421 & 57.1 & $\setminus$ & $\setminus$ & $\setminus$ & $\setminus$ & \textcolor{blue}{109.2} \\
                     &                      & l2d   & 8168.9537 & 34.9 & 4160.1207 & 21.9 & $\setminus$ & $\setminus$ & $\setminus$ & $\setminus$ & \textcolor{blue}{108.8} \\
                     &                      & s2sat & 8268.8350 & 24.3 & 7135.4707 & 19.5 & $\setminus$ & $\setminus$ & $\setminus$ & $\setminus$ & \textcolor{blue}{108.9} \\
                     &                      & s2d   & 5288.4170 & 33.6 & 5309.1103 & 31.7 & $\setminus$ & $\setminus$ & $\setminus$ & $\setminus$ & \textcolor{blue}{109.7} \\
                     & \underline{DAM}      & l2sat & \textcolor{blue}{507.1945} & \textcolor{blue}{55.4} & \textcolor{blue}{208.8668} & \textcolor{blue}{70.8} & \textcolor{blue}{44.2139} & \textcolor{blue}{37.8} & $\setminus$ & $\setminus$ & 110.6 \\
                     &                      & l2d   & \textcolor{blue}{690.1782} & \textcolor{blue}{39.4} & \textcolor{blue}{301.1191} & \textcolor{blue}{66.8} & \textcolor{blue}{96.5603} & \textcolor{blue}{22.5} & $\setminus$ & $\setminus$ & 117.3 \\
                     &                      & s2sat & \textcolor{blue}{740.6881} & \textcolor{blue}{35.2} & \textcolor{blue}{732.4164} & \textcolor{blue}{33.6} & \textcolor{blue}{105.1678} & \textcolor{blue}{24.1} & $\setminus$ & $\setminus$ & 114.4 \\
                     &                      & s2d   & \textcolor{blue}{536.5027} & \textcolor{blue}{51.5} & \textcolor{blue}{616.4043} & \textcolor{blue}{43.9} & \textcolor{blue}{68.1288} & \textcolor{blue}{33.9} & $\setminus$ & $\setminus$ & 114.2 \\
                     & \textbf{\underline{DPCN}}         & l2sat & \textcolor{red}{40.5561} & \textcolor{red}{96.9} & \textcolor{red}{4.8175} & \textcolor{red}{98.0} & \textcolor{red}{0.1172} & \textcolor{red}{99.2} & \textcolor{red}{0.00345} & \textcolor{red}{95.5} & \textcolor{red}{74.39} \\
                     & \textbf{(\underline{Ours})}        & l2d   & \textcolor{red}{15.53} & \textcolor{red}{98.2} & \textcolor{red}{6.4531} & \textcolor{red}{94.0} & \textcolor{red}{0.0412} & \textcolor{red}{99.2} & \textcolor{red}{0.0122} & \textcolor{red}{94.2} & \textcolor{red}{74.63} \\
                     &                      & s2sat & \textcolor{red}{65.373} & \textcolor{red}{90.9} & \textcolor{red}{15.5920} & \textcolor{red}{97.8} & \textcolor{red}{0.1078} & \textcolor{red}{97.4} & \textcolor{red}{0.0055} & \textcolor{red}{93.7} & \textcolor{red}{75.38} \\
                     &                      & s2d   & \textcolor{red}{327.31} & \textcolor{red}{91.3} & \textcolor{red}{14.493} & \textcolor{red}{92.6} & \textcolor{red}{0.2274} & \textcolor{red}{99.3} & \textcolor{red}{0.0070} & \textcolor{red}{93.5} & \textcolor{red}{73.44} \\ \midrule
\multirow{6}{*}{(b)} & \textbf{\textbf{Benchmarks}} & &&&&&&&&&  \\ \cmidrule(lr){2-2}
                     & \underline{QATM}                 & l2d   & 3603.8648 & \textcolor{blue}{37.5} & 4018.3337 & 35.9 & $\setminus$ & $\setminus$ & $\setminus$ & $\setminus$ & \textcolor{blue}{109.2} \\
                     &                      & s2d   & 2808.5308 & 36.3 & 2878.3589 & 31.0 & $\setminus$ & $\setminus$ & $\setminus$ & $\setminus$ & \textcolor{blue}{108.5} \\
                     & \underline{DAM}                   & l2d   & \textcolor{blue}{972.8225} & 30.1 & \textcolor{blue}{588.4123} & \textcolor{blue}{42.2} & \textcolor{blue}{61.3341} & \textcolor{blue}{35.1} & $\setminus$ & $\setminus$ & 113.9 \\
                     &                      & s2d   & \textcolor{blue}{633.2790} & \textcolor{blue}{40.9} & \textcolor{blue}{484.3626} & \textcolor{blue}{49.6} & \textcolor{blue}{85.3438} & \textcolor{blue}{27.4} & $\setminus$ & $\setminus$ & 116.5 \\
                     & \textbf{\underline{DPCN}}        & l2d   & \textcolor{red}{8.0043} & \textcolor{red}{96.2} & \textcolor{red}{102.359} & \textcolor{red}{89.2} & \textcolor{red}{0.0059} & \textcolor{red}{99.7} & \textcolor{red}{0.0005} & \textcolor{red}{99.7} & \textcolor{red}{75.34}\\
                     & \textbf{(\underline{Ours})}                & s2d   & \textcolor{red}{88.7428} & \textcolor{red}{91.6} & \textcolor{red}{61.0860} & \textcolor{red}{90.6} & \textcolor{red}{0.7634} & \textcolor{red}{99.4} & \textcolor{red}{0.0035} & \textcolor{red}{95.0} & \textcolor{red}{75.96} \\
\bottomrule[1pt]
\end{tabular}
}
\vspace{-10pt}
    \end{table}

    Table \ref{results: Aero-Ground Dataset} shows validation results on scene (a) and (b) with the threshold error of $10 \:pixels$ for translation, $1\degree$ for rotation and $0.2 \times$ for scale. Acknowledging that each ground-image is generated with a scale of $0.1 \:meters\:per\:pixel(MPP)$ in the Aero-Ground Dataset, the threshold error of $10\:pixels$ for translation could be transformed to $1\:meter$ in the real world. The results prove that when estimating 4-DoF poses across real-world sensors, our approach is the first to finish this job with an accuracy of at least $89.22\%$ when considering error lower than $1\:meter$. Even when we relax the conditions and provide initials to the rest of the benchmarks, our approach still outperforms the rest of them in 2-DoF(QATM) and 3-DoF(DAM). Additional demonstrations of the experiment comparing with conventional phase correlation is shown in appendix \ref{appendix: Additional Demonstration}.

\begin{figure}[h]
 \vspace{-0.7cm}
    \captionsetup[subfigure]{justification=centering}
    \centering
  \subfloat[Estimation of $Acc_{x_{0\sim19}}$ \label{Generalization AG_x}]{%
        \includegraphics[width=0.33\linewidth]{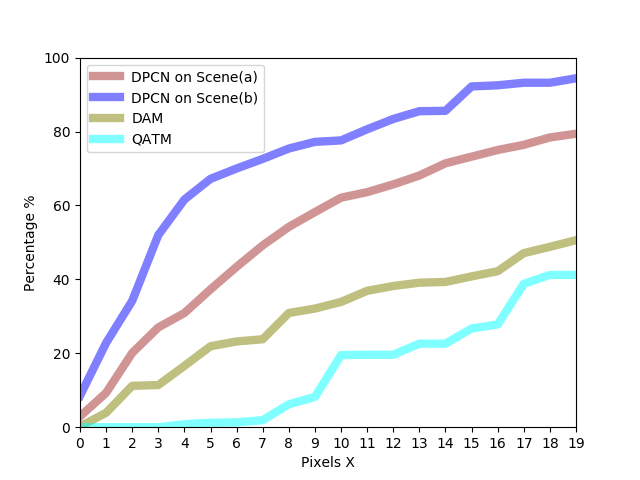}}
  \subfloat[Estimation of $Acc_{y_{0\sim19}}$\label{Generalization AG_y}]{%
        \includegraphics[width=0.33\linewidth]{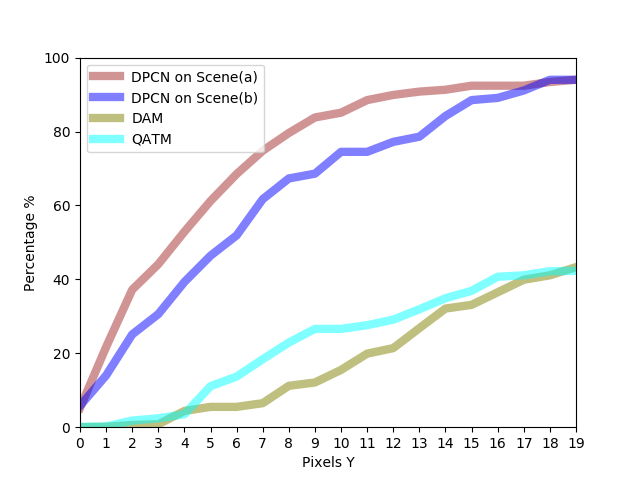}}
  \subfloat[Estimation of $Acc_{rot_{0\sim19}}$\label{Generalization AG_rot}]{%
        \includegraphics[width=0.33\linewidth]{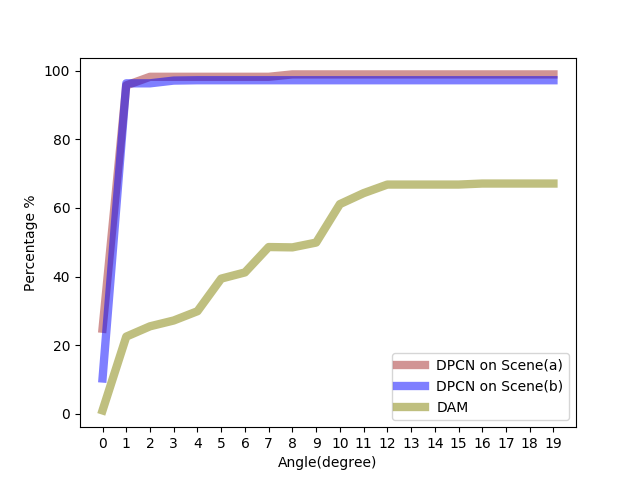}}
    \vspace{-0.2cm}
  \caption{Transformation estimation in generalization on scene (c).}
  \label{fig:AG Generalization}
  \vspace{-0.5cm}
\end{figure}

    To evaluate the generalizing capability of our approach, we conduct experiments on scene (c) with DPCN models trained on scene(a) and (b) and DAM model trained on scene(c). The results shown in Figure \ref{fig:AG Generalization} and Table \ref{results:generalizationAG}(Appendix \ref{appendix: Elaboration on Translation Estimation}) prove that with the types of source sensors given and fixed, our approach is capable of estimating poses regardless of scene changes and illumination changes with similar accuracy and still outperformed DAM which is specifically trained on (c). Therefore, the robustness of our approach in the real-world application is well documented.

\subsection{Application in satellite map based localization}
\label{sec:mcl}

\begin{wrapfigure}{r}{0.5\textwidth}
\vspace{-15pt}
  \begin{center}
    \includegraphics[width=0.5\textwidth]{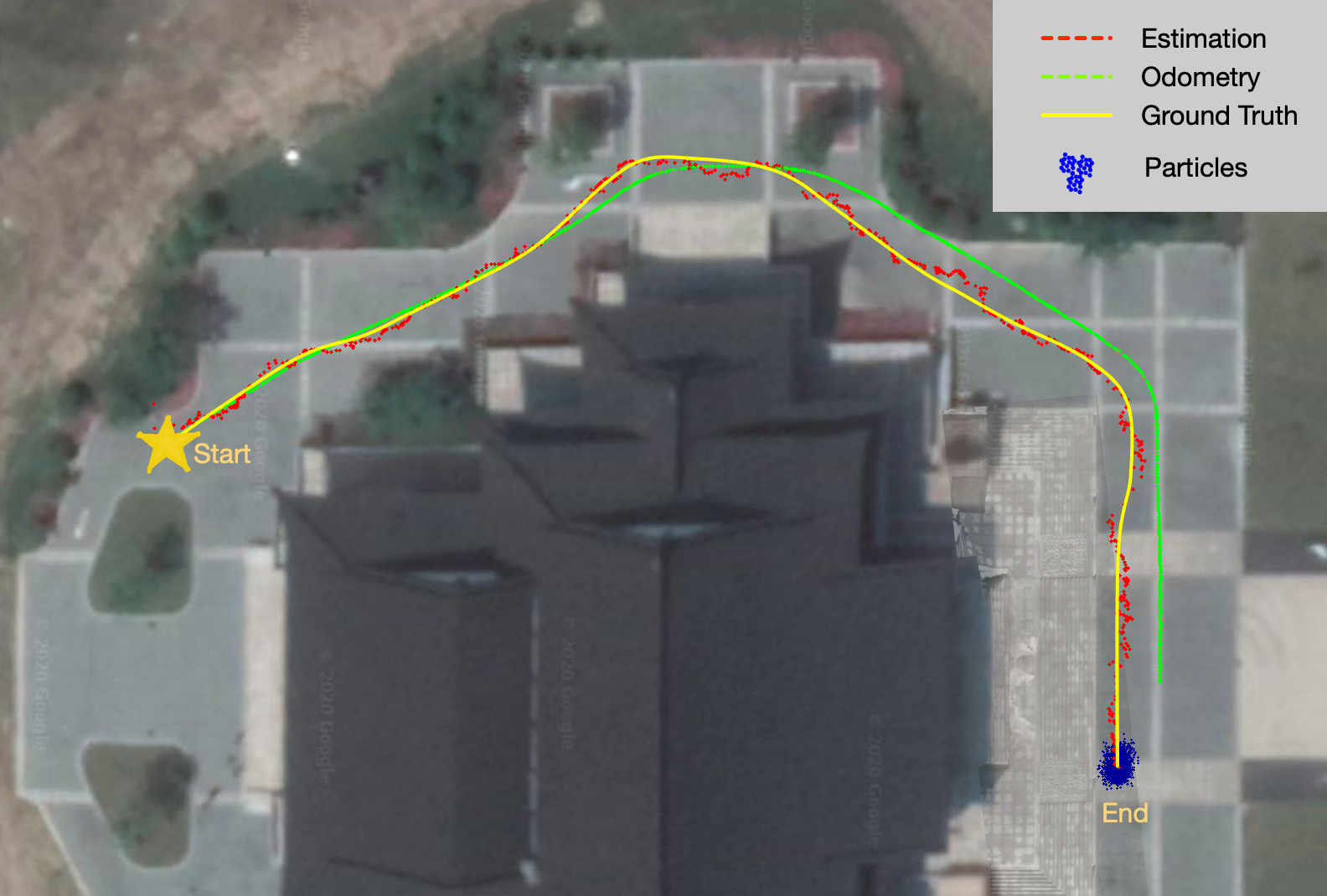}

  \end{center}
  \vspace{-8pt}
  \caption{Application in Mont Carlo Localization.}
  \vspace{-5pt}
  \label{fig:MCL}
\end{wrapfigure}
In this section, we demonstrate the application in localization by introducing our approach to Monte Carlo Localization. It proves that with corresponding maps as the output, our model is capable of real-time air-ground localization, e.g. scene(a). A demonstration is shown in Figure \ref{fig:MCL}, where the green dashed line is the odometry estimated by ``VINS" through a stereo camera, the red dashed line is generated by MCL by matching 4-DoF poses of LiDAR intensity map and satellite ,ap, and the yellow line is the ground truth. As the cumulative error of the odometry gradually increases, the corrective effect of our method is sufficiently demonstrated.


\section{Conclusion}
\label{sec:conclusion}
We present an approach for precise multi-sensor pose matching which greatly fascinate multi-agent collaborative exploration. We achieve this by training pairs of individual U-Nets with end-to-end poses to learn representations of heterogeneous images to be recognized by differentiable phase correlation(DFT+LPT+Correlation). We show that by training the network end-to-end with a fully differentiable pipeline, the network is easy and fast to be trained, precise in matching and capable of running in real-time. We also show that with every estimation analytical, the network is completely interpretive and has the capability of generalization.


\clearpage
\acknowledgments{This project is supported and funded by the National Key R\&D Program of China(2018AAA0102703), the National Nature Science Foundation of China(61903332), and the Science and Technology on Space Intelligent Control Laboratory(Grant No.HTKJ2019KL502002). We give sincere appreciations to all the reviewers.}


\bibliography{LPTM_new}  

\clearpage

\appendix

\section{Network Structure and Experimental Setup}
\label{appendix: Network Structure and Experimental Setup}

\subsection{Network Structure}
\label{appendix sub: Network Structure}
We train the DPCN network with the input of pairs of heterogeneous images and the ground truth of their relative pose. There are two phases here in the training. \textbf{Phase 1}: Rotation and scale changes are trained and estimated in this phase. The pair of images go through a pair of U-Nets and their outputs go through the DFT layer for translation reduction. The Log Polar Transform remaps the DFT's output to the log polar coordinate so that rotation and scale variances are then shown in columns and rows. Afterward, phase correlation is able to estimate the rotation and scale with a relative power spectrum as the output. Finally, we supervised the U-Net using Cross-Entropy Loss of both rotation, scale, and their ground truth. \textbf{Phase 2}: Translation of the image is trained and estimated in this phase. The source image of the image pair is rotated and scaled by the result in phase 1 and the new image pair(original template image and the new source image) go through a new pair of codec. The translation is then estimated by phase correlation and therefore the codec is supervised using Cross-Entropy Loss and L1 Loss of estimated transformation and their ground truth.

\begin{figure}[ht]
    \centering
    \includegraphics[width=\linewidth]{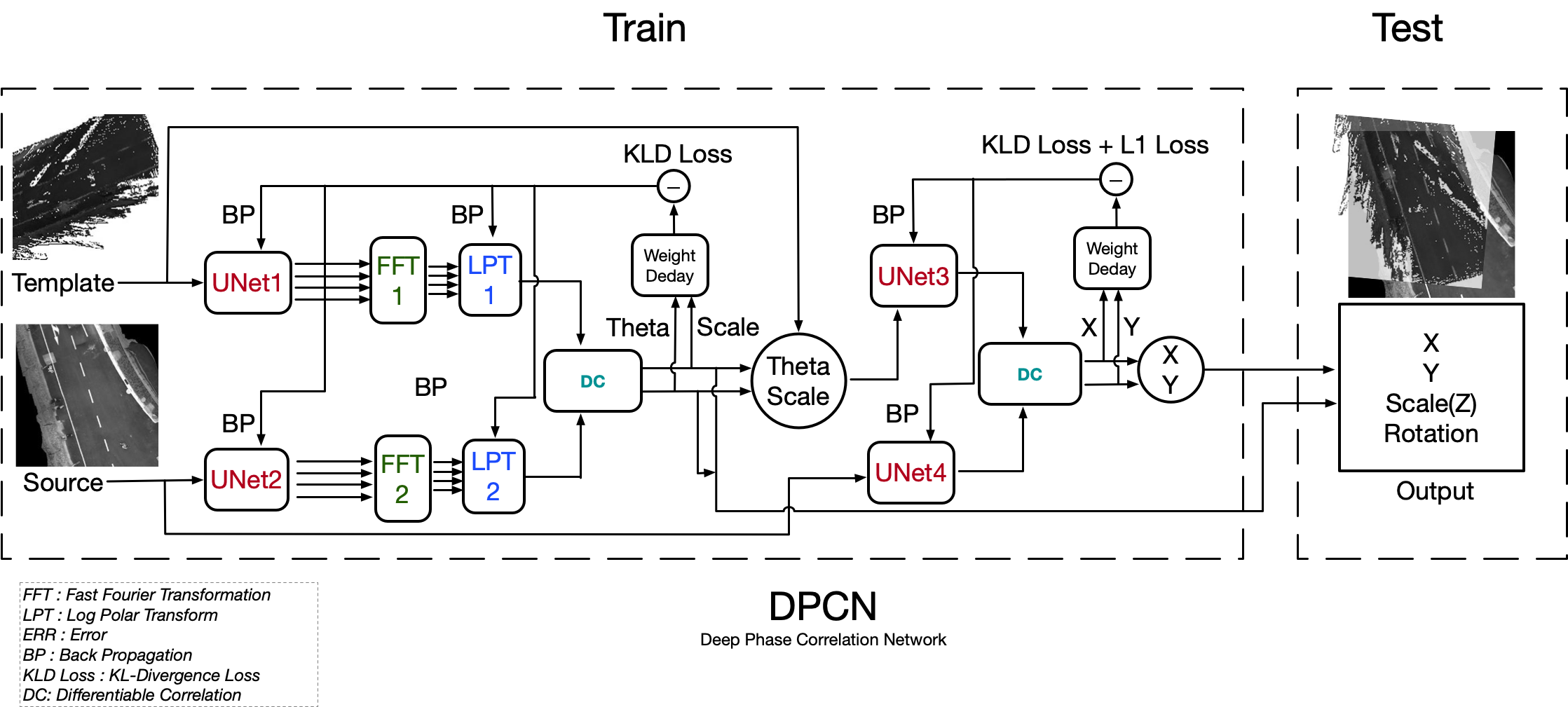}
    \caption{The network structure of DPCN}
\label{fig:net structure}
\vspace{-0.4cm}
\end{figure}

\subsection{Experimental Setup}
\label{appendix sub: Experimental Setup}

\textbf{Hardware: } All models for the full method and their comparison models were trained on a single server with CPU of Intel i9-9900X @ 3.5GHz x 20 and GPU of RTX 2080ti x 4. The validation experiments are all conducted on a single desktop computer(AMD Ryzen 3700X @ 3.7GHz x8) with the GPU of an RTX 2060super. The Aero-Ground Dataset is record by a ground robot and a DJI M100 drone. The ground robot is equipped with three 32-wire Velodyne LiDAR, four pairs of stereo cameras by Flir and one RTK dGPS provided by QX. The DJI drone records videos by one pair of Intel Realsense D435i stereo camera with one facing forward and one downward.

\textbf{Software:} The birds-eye global map from the drone used in the training and validation is constructed by full-licensed software \textsc{Metashape} with video clips from the Aero-Ground dataset. The local map of both LiDAR intensity style and stereo style from the ground robot is constructed by \textsc{Elevation Map}\cite{Pan2019}. All of the maps constructed above and the satellite map obtained from \textsc{Google Maps} have the resolution of 0.1 MPP.

\clearpage

\section{Additional Demonstration}
\label{appendix: Additional Demonstration}

\begin{figure}[ht]
\centering
\includegraphics[width=\linewidth]{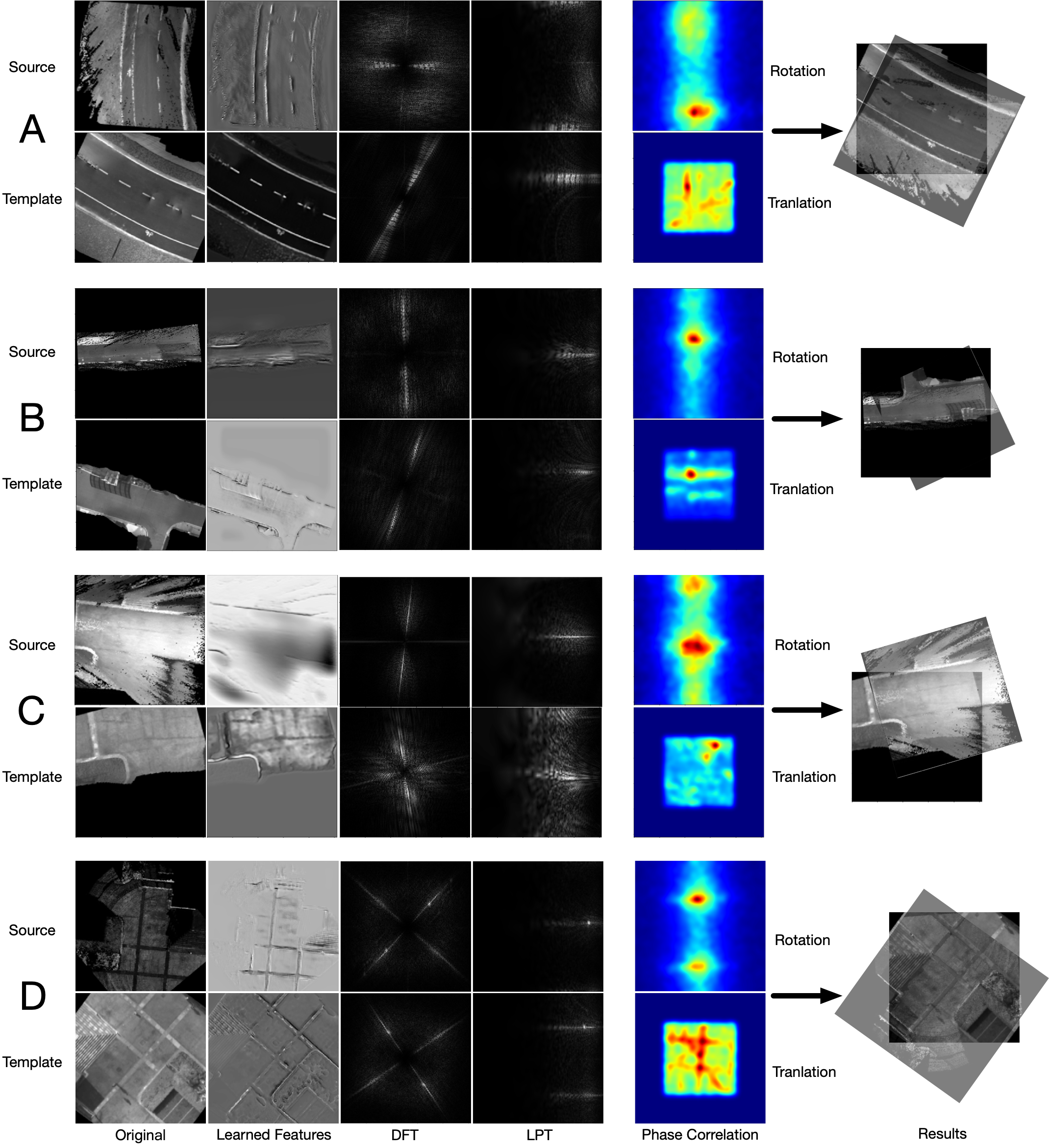}
\caption{Additional four demonstrations matching heterogeneous images pairs from Aero-Ground Dataset. The respective comparisons are shown in subsection \ref{subsec:PC}. }
\label{fig: additional demo}
\vspace{-0.5cm}
\end{figure}

\begin{figure}[ht]
\centering
\includegraphics[width=0.85\linewidth]{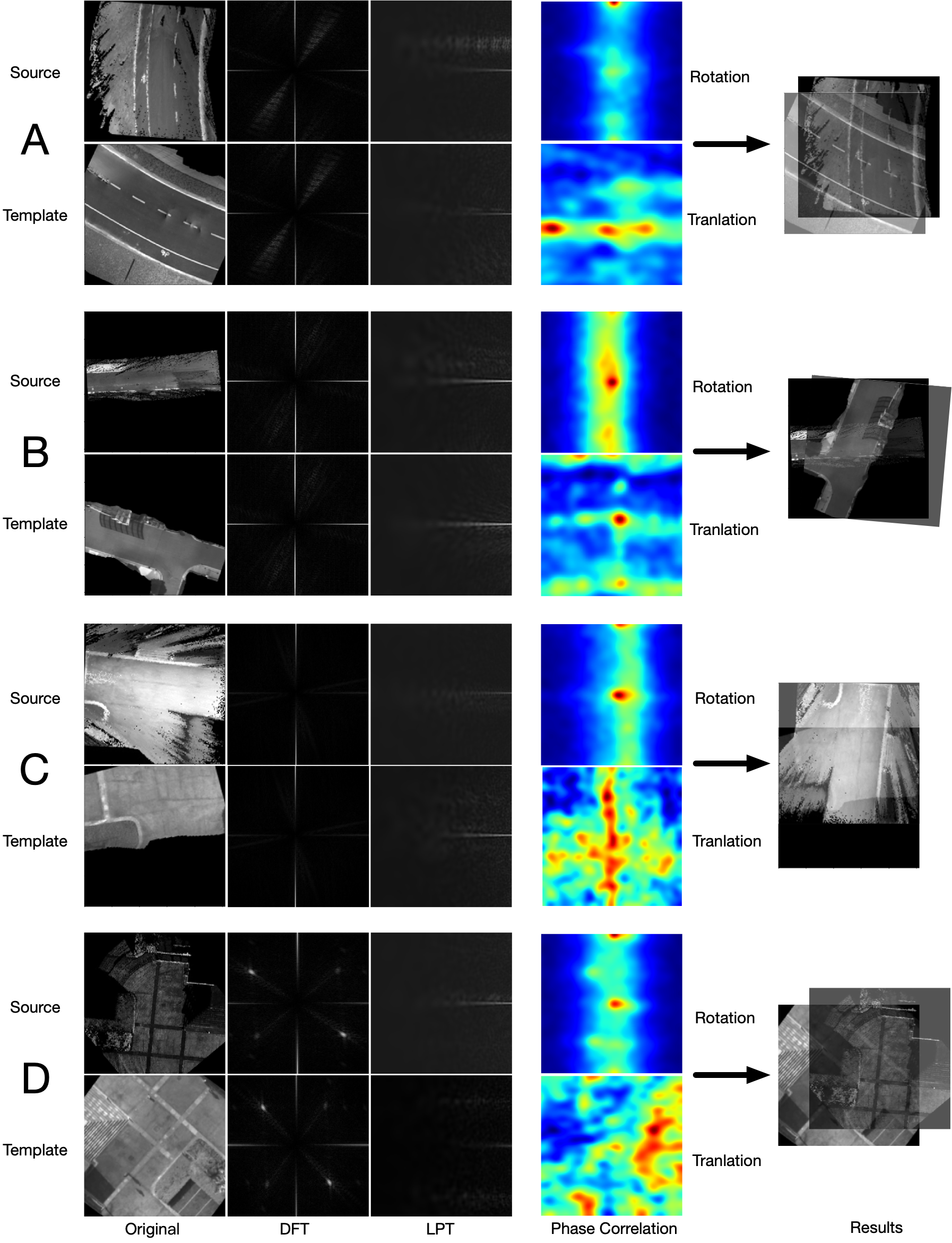}
\caption{Comparison using conventional phase correlation to match heterogeneous images pairs from subsection \ref{subsec:DPCN}.}
\label{fig: additional compare}
\vspace{-0.5cm}
\end{figure}

Demonstrations of comparing DPCN and conventional phase correlation on real-world dataset Aero-Ground Dataset is shown in Figure \ref{fig: additional demo} and Figure \ref{fig: additional compare}. They show that when matching heterogeneous images from different sensors, the trainable feature extractors in DPCN play important roles in outperforming the conventional phase correlation. However, the core of the entire DPCN is the differentiable DFT, LPT and phase correlation that could back propagate losses and eventually train the feature extractors. They also prove that by learning features through the supervision of the end-to-end poses, the approach is capable of reducing hollows, rejecting noises and assimilate different styles to estimate the 4-DoF relative pose.

\subsection{DPCN Results}
\label{subsec:DPCN}

\clearpage

\subsection{Conventional Phase Correlation Results}
\label{subsec:PC}

\clearpage

\section{Elaboration on Estimation}
\label{appendix: Elaboration on Translation Estimation}
\begin{figure}[h]
    \captionsetup[subfigure]{justification=centering}
    \centering
  \subfloat[Estimation of $x$ in Homogeneous dataset\label{Homo_x}]{%
      \includegraphics[width=0.5\linewidth]{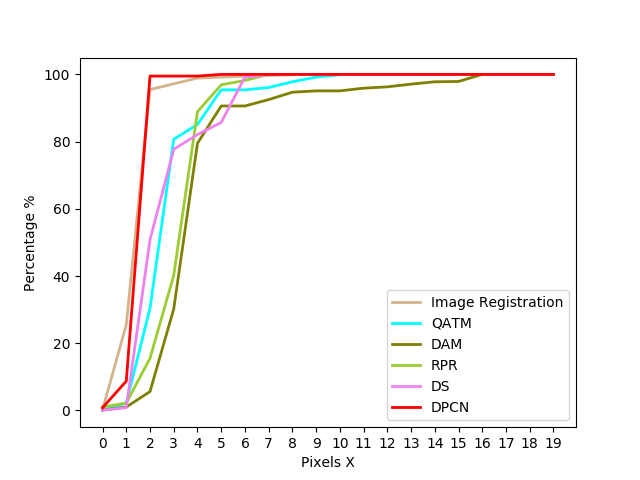}}
  \subfloat[Estimation of $y$ in Homogeneous dataset\label{Homo_y}]{%
        \includegraphics[width=0.5\linewidth]{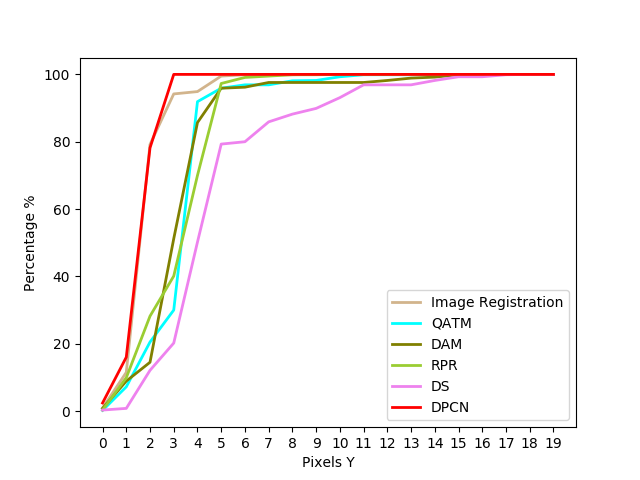}}
        \\
  \subfloat[Estimation of $x$ in Heterogeneous dataset\label{Hetero_x}]{%
        \includegraphics[width=0.5\linewidth]{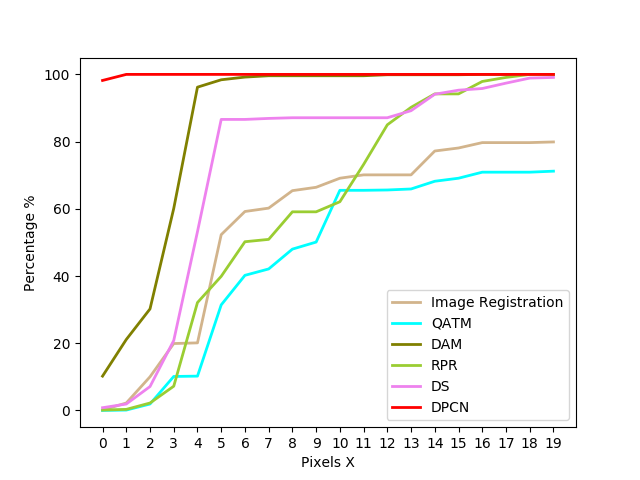}}
  \subfloat[Estimation of $y$ in Heterogeneous dataset\label{Hetero_y}]{%
        \includegraphics[width=0.5\linewidth]{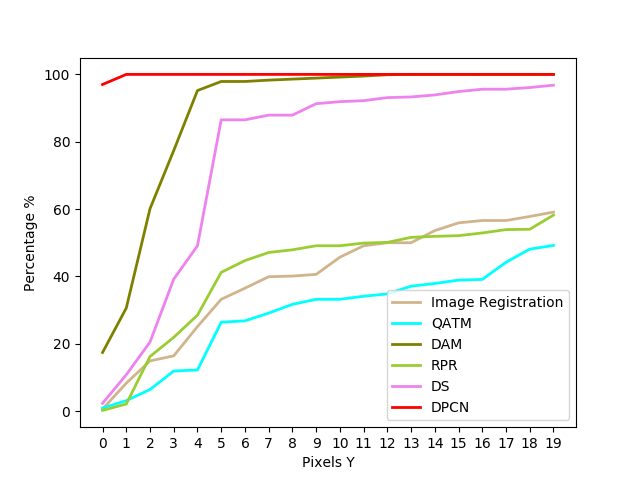}}
        \\
  \subfloat[Estimation of $x$ in Dynamic Obstacle dataset\label{Dynamic_x}]{%
        \includegraphics[width=0.5\linewidth]{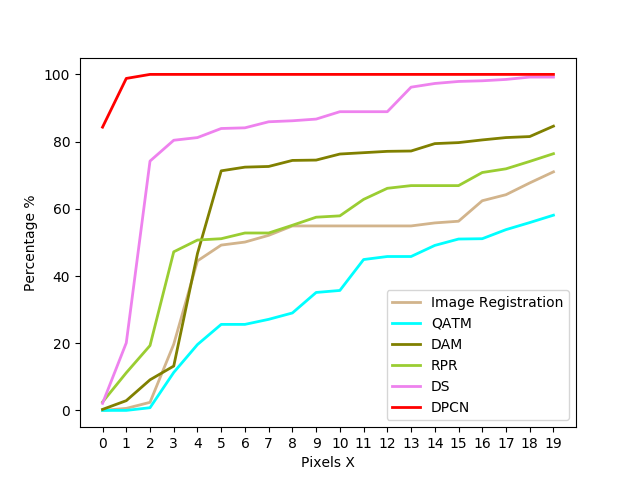}}
  \subfloat[Estimation of $y$ in Dynamic Obstacle dataset\label{Dynamic_y}]{%
        \includegraphics[width=0.5\linewidth]{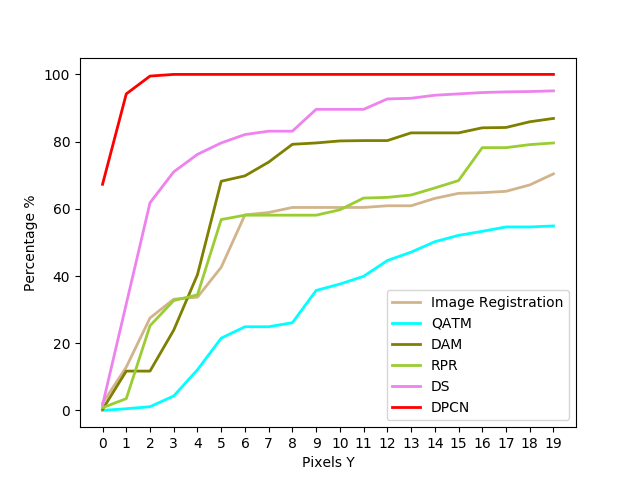}}
  \caption{$Acc_{0\:to\:19}$ of translation estimation in simulation dataset. }
  \label{Acc simulation}
\end{figure}

The threshold of estimation in experiments in \ref{sec:result} is elaborated in this section by the means of graphs. Figure \ref{Acc simulation} shows the $Acc_{0\:to\:19}$ of translation estimation in simulation dataset, figure \ref{Acc AG1} shows the $Acc_{0\:to\:19}$ of translation estimation in Aero-Ground dataset, Figure \ref{Acc Generalization} shows the $Acc_{0\:to\:19}$ of translation estimation in simulation dataset, and Table \ref{results:generalizationAG} show the exact error of generalization experiments on Aero-Ground Dataset.

 \begin{figure}[h]
    \captionsetup[subfigure]{justification=centering}
    \centering
  \subfloat[Estimation of $x$ in ``LiDAR to Drone" scene(a)\label{qsdjt_l2a_x}]{%
      \includegraphics[width=0.5\linewidth]{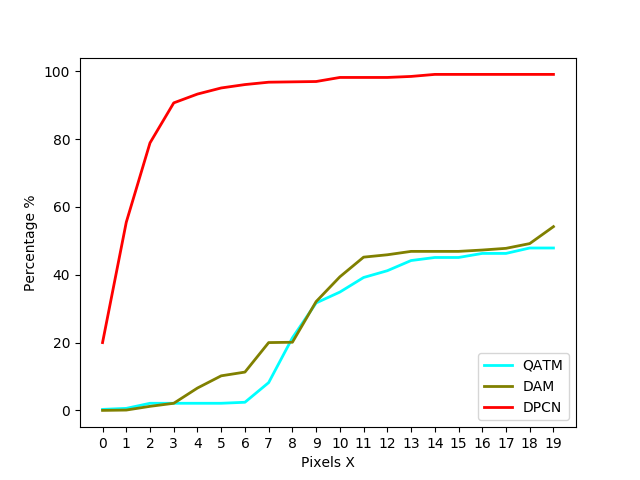}}
  \subfloat[Estimation of $y$ in ``LiDAR to Drone" scene(a)\label{qsdjt_l2a_y}]{%
        \includegraphics[width=0.5\linewidth]{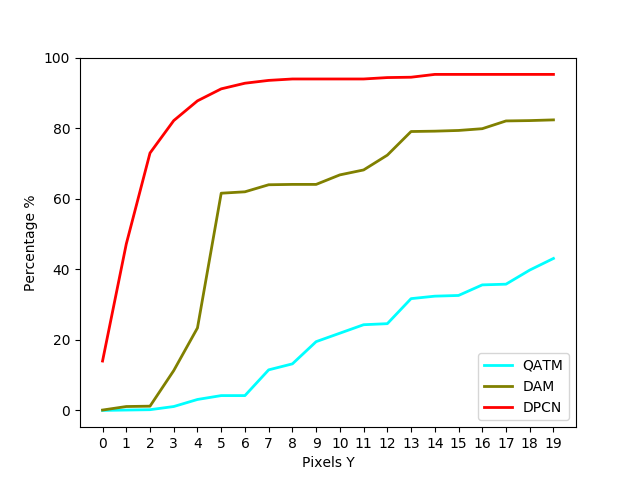}}
        \\
  \subfloat[Estimation of $x$ in ``LiDAR to Satellite" scene(a)\label{qsdjt_l2s_x}]{%
        \includegraphics[width=0.5\linewidth]{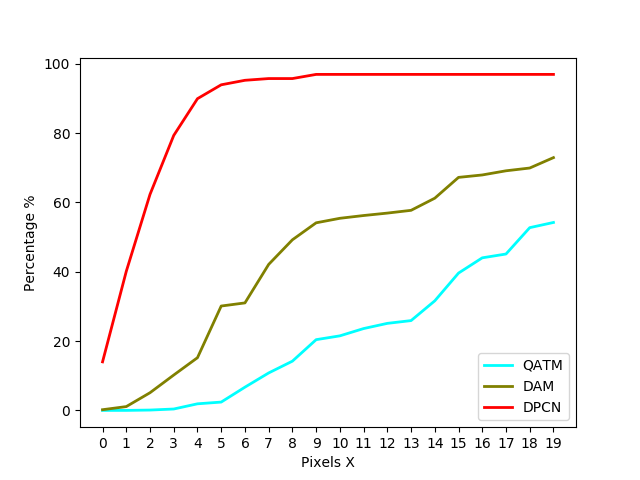}}
  \subfloat[Estimation of $y$ in ``LiDAR to Satellite" scene(a)\label{qsdjt_l2s_y}]{%
        \includegraphics[width=0.5\linewidth]{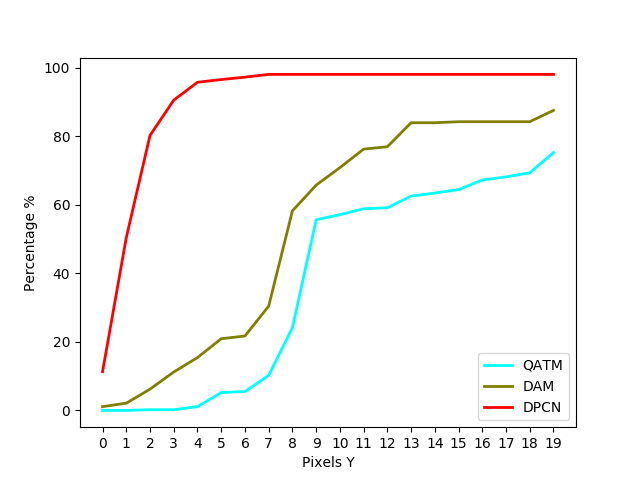}}
        \\
  \subfloat[Estimation of $x$ in ``Stereo to Drone" scene(a)\label{qsdjt_s2a_x}]{%
      \includegraphics[width=0.5\linewidth]{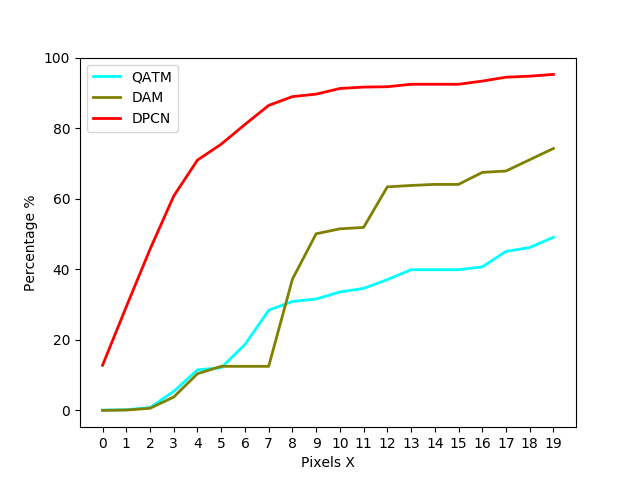}}
  \subfloat[Estimation of $y$ in ``Stereo to Drone" scene(a)\label{qsdjt_s2a_y}]{%
        \includegraphics[width=0.5\linewidth]{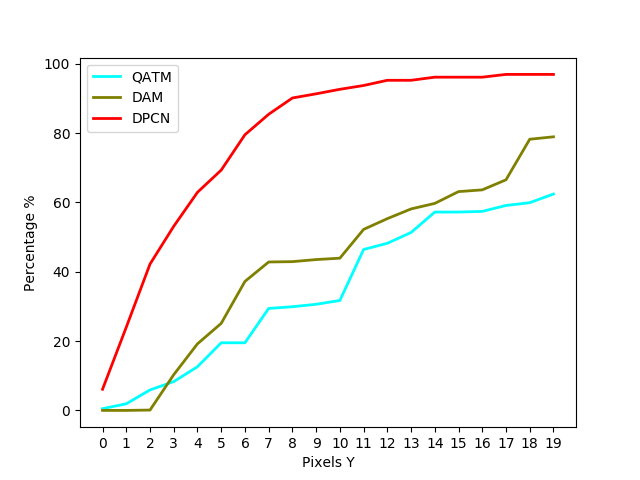}}
  \caption{$Acc_{0\:to\:19}$ of translation estimation in Aero-Ground dataset. }
  \label{Acc AG1}
\end{figure}

 \begin{figure}[h]
 \ContinuedFloat
    \captionsetup[subfigure]{justification=centering}
    \centering
  \subfloat[Estimation of $x$ in ``Stereo to Satellite" scene(a)\label{qsdjt_s2s_x}]{%
        \includegraphics[width=0.5\linewidth]{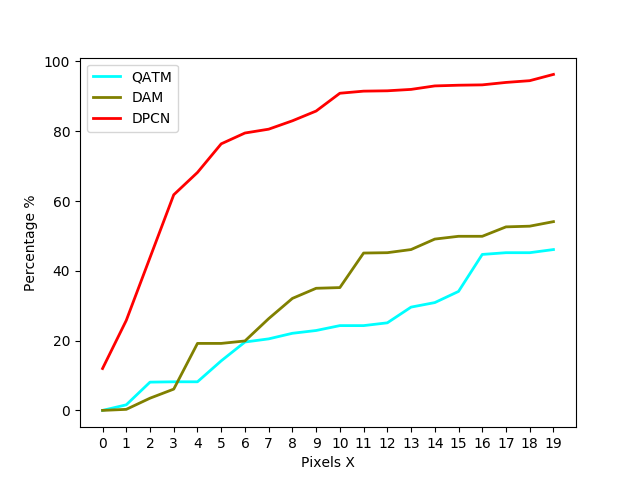}}
  \subfloat[Estimation of $y$ in ``Stereo to Satellite" scene(a)\label{qsdjt_s2s_y}]{%
        \includegraphics[width=0.5\linewidth]{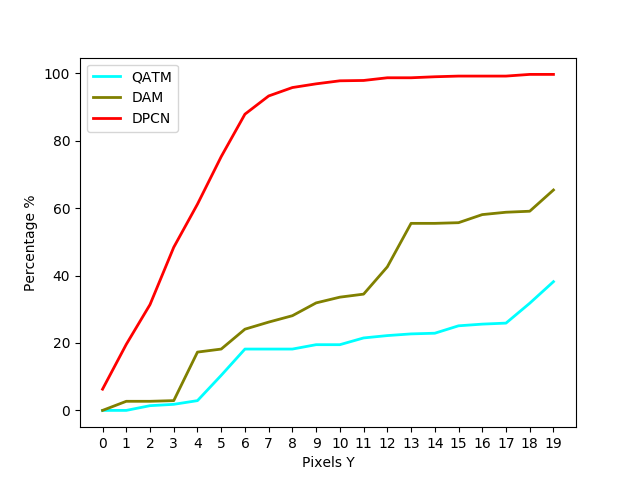}}
        \\
  \subfloat[Estimation of $x$ in ``LiDAR to Drone" scene(b)\label{gym_l2a_x}]{%
        \includegraphics[width=0.5\linewidth]{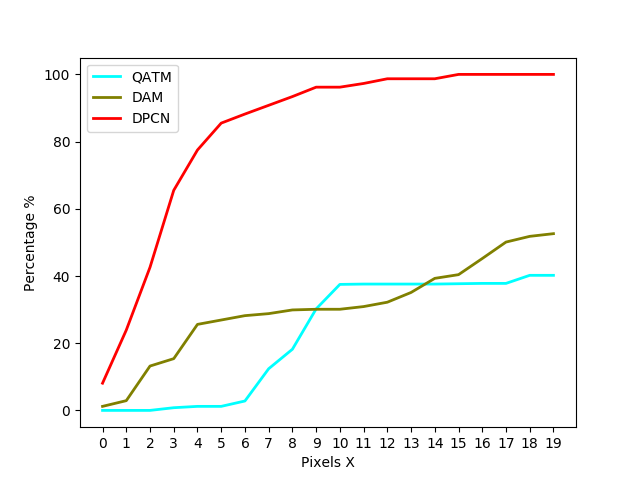}}
  \subfloat[Estimation of $y$ in ``LiDAR to Drone" scene(b)\label{gym_l2a_y}]{%
        \includegraphics[width=0.5\linewidth]{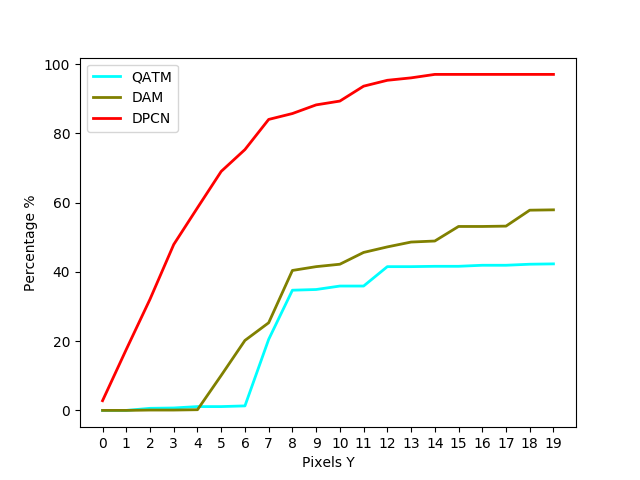}}
        \\
  \subfloat[Estimation of $x$ in ``Stereo to Drone" scene(b)\label{gym_s2a_x}]{%
        \includegraphics[width=0.5\linewidth]{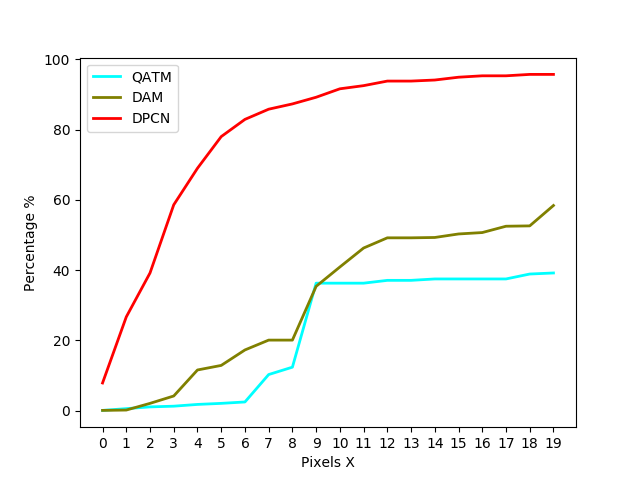}}
  \subfloat[Estimation of $y$ in ``Stereo to Drone" scene(b)\label{gym_s2a_y}]{%
        \includegraphics[width=0.5\linewidth]{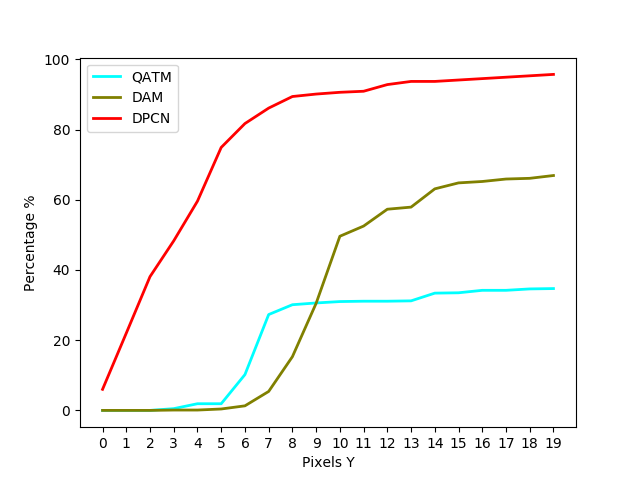}}
  \caption{$Acc_{0\:to\:19}$ of translation estimation in Aero-Ground dataset. }
  \label{Acc AG2}
\end{figure}

 \begin{figure}[ht]
    \captionsetup[subfigure]{justification=centering}
    \centering
  \subfloat[Estimation of $x$ in simulation generalization\label{Generalization Sim_x}]{%
      \includegraphics[width=0.5\linewidth]{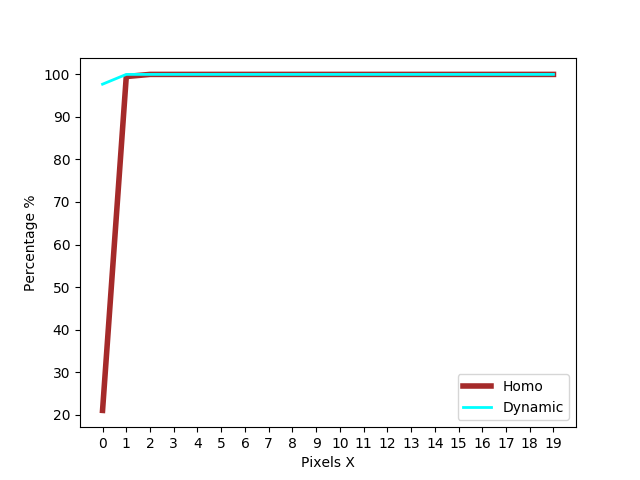}}
  \subfloat[Estimation of $y$ in simulation generalization\label{Generalization Sim_y}]{%
        \includegraphics[width=0.5\linewidth]{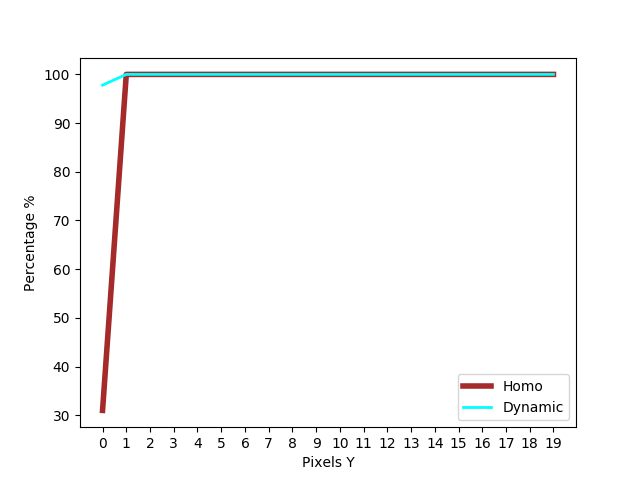}}

  \caption{$Acc_{0\:to\:19}$ of translation estimation in generalization. }
  \label{Acc Generalization}
\end{figure}

\begin{table}[ht]
\renewcommand\arraystretch{1.4}
\caption{Results of generalization experiments with Aero-Ground Dataset. Experiments are conducted with the input type of stereo camera and drone's birds-eye (AKA ``s2d" in Table \ref{results: Aero-Ground Dataset}), therefore, the model applied in these experiments are trained on the ``s2d" dataset in scene (a) and (b). For generalization, we choose the threshold error of $15 \:pixels$ for translation, $1\degree$ for rotation and $0.2\times$ for scale. More threshold is elaborated in Figure \ref{fig:AG Generalization}}
\centering
\begin{small}
\resizebox{\textwidth}{!}{
\begin{tabular}{c ccccccccc ccccccccc }
\toprule[1pt]
Model & $E_x$ & $Acc_{x_{15}}(\%)$ & $E_y$ & $Acc_{y_{15}}(\%)$ & $E_{rot}$ & $Acc_{rot_{1}}(\%)$ & $E_{scale}$ & $Acc_{scale_{0.2}}(\%)$ \\ \midrule
DPCN in (a)       &  232.4638  &  73.2   &  29.9253   &  92.4   &  89.0943  &  95.7   &  0.0084  & 95.0\\
DPCN in (b)       &  31.2071  &  92.2   &  138.5449   &  88.5   &  2.8793  &  96.3   &  0.0153  & 93.3\\
DAM       &  602.8490  &  40.8   &  720.9244   &  33.1   &  88.7239  &  22.5   &  0.0153  & 93.3\\
QATM       &  3922.6715  &  26.7   &  1103.6291   &  36.9   &  $\setminus$  &  $\setminus$   &  $\setminus$  & $\setminus$\\
\bottomrule[1pt]
\end{tabular}
}
\end{small}
\label{results:generalizationAG}
\end{table}

\clearpage

\end{document}